\documentclass[letterpaper]{article} 
\usepackage{aaai24}  
\usepackage{times}  
\usepackage{helvet}  
\usepackage{courier}  
\usepackage[hyphens]{url}  
\usepackage{graphicx} 
\usepackage{natbib}  
\usepackage{caption} 
\usepackage{algorithm}
\usepackage{algorithmic}
\usepackage{algorithm}
\usepackage{algorithmic}
\usepackage{amsmath}
\usepackage{amssymb}
\usepackage{multirow}
\usepackage{makecell}
\usepackage{color}
\usepackage[title]{appendix}

\usepackage{newfloat}
\usepackage{listings}
\DeclareCaptionStyle{ruled}{labelfont=normalfont,labelsep=colon,strut=off} 
\floatstyle{ruled}
\newfloat{listing}{tb}{lst}{}
\floatname{listing}{Listing}
\title{Enhancing the Spatial Awareness Capability of Multi-Modal \\Large Language Model}
\author{
    Yongqiang Zhao\textsuperscript{1},
    Zhenyu Li\textsuperscript{2},
    Zhi Jin\textsuperscript{1}\equalcontrib,
    Feng Zhang\textsuperscript{2},
    Haiyan Zhao\textsuperscript{1},
    Chengfeng Dou\textsuperscript{1},\\
    Zhengwei Tao\textsuperscript{1},
    Xinhai Xu\textsuperscript{2},
    Donghong Liu\textsuperscript{2}\equalcontrib
}
\affiliations{
    \textsuperscript{\rm 1}School of Computer Science, Peking University, Beijing, China\\
    \textsuperscript{\rm 2}Academy of Military Sciences, Beijing, China\\

}

\usepackage{bibentry}

\begin{document}

\maketitle
\begin{abstract}
The Multi-Modal Large Language Model (MLLM) refers to an extension of the Large Language Model (LLM) equipped with the capability to receive and infer multi-modal data.
Spatial awareness stands as one of the crucial abilities of MLLM, encompassing diverse skills related to understanding spatial relationships among objects and between objects and the scene area. Industries such as autonomous driving, smart healthcare, robotics, virtual, and augmented reality heavily demand MLLM's spatial awareness capabilities. However, there exists a noticeable gap between the current spatial awareness capabilities of MLLM and the requirements set by human needs.
To address this issue, this paper proposes using more precise spatial position information between objects to guide MLLM in providing more accurate responses to user-related inquiries. Specifically, for a particular multi-modal task, we utilize algorithms for acquiring geometric spatial information and scene graphs to obtain relevant geometric spatial information and scene details of objects involved in the query. Subsequently, based on this information, we direct MLLM to address spatial awareness-related queries posed by the user.
Extensive experiments were conducted in benchmarks such as MME, MM-Vet, and other multi-modal large language models. The experimental results thoroughly confirm the efficacy of the proposed method in enhancing the spatial awareness tasks and associated tasks of MLLM.
\end{abstract}

\section{Introduction}
Recently, the Multi-Modal Large Language Model (MLLM)~\cite{wang2023large, huang2023chatgpt, zhang2023speechgpt} has emerged as a hot research area. It utilizes a powerful Large Language Model (LLM)~\cite{kasneci2023chatgpt, liu2023summary} as a cognitive engine, focusing on executing multi-modal tasks, which holds significant importance in advancing research and application in multi-modal understanding. MLLM finds wide applications across various fields~\cite{yin2023survey, chang2023two, li2023videochat}, including autonomous driving, smart healthcare, robotics, e-commerce, virtual, and augmented reality. It significantly enhances the comprehension and processing of multi-modal data and stands as a crucial pathway in achieving artificial intelligence.

\begin{figure}[t]
\centering
\includegraphics[width=1.0\columnwidth]{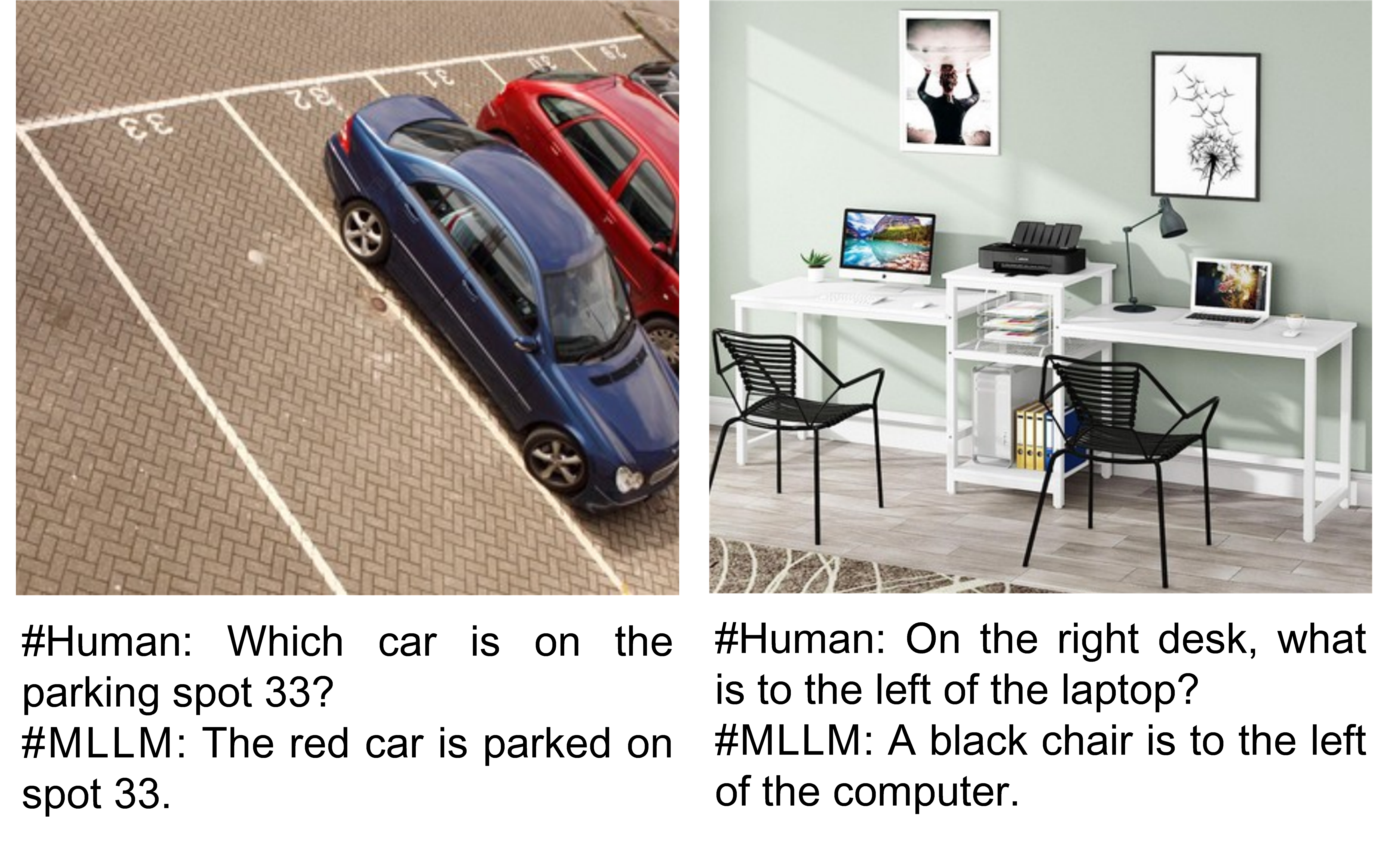} 
\caption{Instance of a multi-modal large language model in spatial awareness task}
\label{fig:1}
\end{figure}

Among its capabilities, perception stands as one of the fundamental aspects of MLLM, signifying its ability to accurately acquire external information. Within perception, spatial awareness particularly holds significance as it encompasses various abilities related to understanding spatial relationships between objects or between objects and the surrounding scene area. In many application scenarios of MLLM, spatial awareness demands stringent precision. For instance, in the domain of autonomous driving, precise localization of objects such as vehicles, pedestrians, traffic signs, road markings, parking spaces, etc., is essential to ensure safe driving and prevent accidents. Similarly, in smart healthcare, precise localization of structures like tumors, lesions, organs inside a patient's body is crucial for accurate diagnosis and treatment plans. However, the current performance of MLLM in spatial awareness still significantly lags behind human requirements. As depicted in Figure~\ref{fig:1}, existing MLLMs struggle to accurately determine spatial relationships between objects, such as the spatial relationship between the red car and parking spot 33 (left), and between the desk, laptop, and table lamp (right).

This research aims to enhance the spatial awareness capability of MLLM. One of the most direct approaches involves providing MLLM with more precise spatial awareness information, using this exact information as input to guide the generation of results by MLLM. Thus, we propose leveraging pretrained smaller models to offer spatial position relationships between target objects and, subsequently, use this information to guide MLLM in addressing user-related queries. Specifically, for specific multimodal tasks, we utilize pretrained object detection algorithms~\cite{ren2017faster, liu2020deep, zou2023object} and scene graph generation algorithms~\cite{yang2022panoptic, chang2021comprehensive, cong2023reltr} to acquire geometric spatial information and scene details pertinent to the query. Based on this information, we direct MLLM to address spatial awareness-related user queries.

The main contributions of this paper can be summarized in two points:
\begin{itemize}
\item Proposing a novel approach to enhance MLLM's spatial awareness capability by using pretrained smaller models to provide geometric spatial information and high-level semantic details between objects, thereby guiding MLLM in generating more accurate results. To our knowledge, this is the first work to combine pretrained smaller models with MLLM to enhance its capabilities.
\item Conducting extensive experiments on benchmarks like MME, MM-Vet, and other MLLM benchmarks. The experimental results thoroughly confirm the effectiveness of the proposed research method in significantly enhancing MLLM's performance in spatial awareness and associated tasks.
\end{itemize}

\section{Related Work}
The Multi-Modal Large Language Model (MLLM)~\cite{dong2022survey, lu2023chameleon, zhang2023multimodal, wu2023visual, liang2023taskmatrix} leverages a powerful Large Language Model (LLM) as its cognitive engine to handle various multi-modal tasks. In many application scenarios, collecting diverse types of information through multiple input channels is necessary to construct specific task models. For instance, in autonomous driving, a vehicle needs to process data from both cameras and LiDAR to make effective decisions in complex driving environments. MLLM excels in understanding and processing a variety of information from different modalities, thus playing a vital role in advancing research and application of multi-modal comprehension.

Current research on MLLM can be broadly categorized into two types: the first type comprises models composed of LLM and visual encoders, such as LLaVa~\cite{liu2023visual}, MiniGPT-4~\cite{zhu2023minigpt}, mPLUG-Owl~\cite{ye2023mplug}. These models aim to achieve the multi-modal understanding effect presented in the GPT4 technical report with minimal additional training based on existing LLM. The second type involves models composed of LLM and various small multi-modal models, such as HuggingGPT~\cite{shen2023hugginggpt}, MOSS, CompeGPT~\cite{zhao2023enhancing}. These models utilize LLM as a control and organization center, calling different small models to accomplish distinct multi-modal tasks and subsequently consolidating user responses.

This study is rooted in the first type of MLLM. These models share similarities in their structure and training strategies, with differences mainly apparent in whether the LLM and visual encoders are frozen. For instance, LLaVA, MiniGPT-4 freeze the basic visual encoders, while mPLUG-Owl releases the visual encoders. Related research has also demonstrated that pre-training on image-text is critical for establishing connections between image and text. Current work primarily focuses on how to enhance the overall model performance using MLLM's inherent capabilities. Nevertheless, "there is no gold standard, and no one is perfect"; the present MLLM struggles to effectively accomplish all multi-modal tasks with its intrinsic abilities alone. Hence, this study considers utilizing more precise spatial awareness information acquired from external models to guide the result generation of MLLM.

\section{Method}
In this section, we first present an overview of the proposed method and then provide detailed information about its main components.

\subsection{Overview}
In response to the given multi-modal request, we initially employ pre-trained object detection algorithms and scene graph generation algorithms to acquire the geometric spatial position information and scene graph details related to spatial awareness queries. Subsequently, we guide the Multi-Modal Large Language Model (MLLM) to address user-related queries based on this information.

\begin{figure*}
\centering
\includegraphics{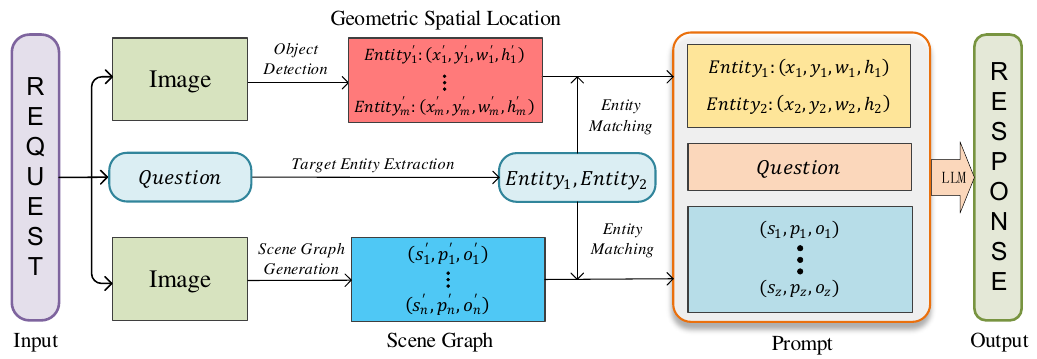}
\caption{The overview of our method.}
\label{fig:2}
\end{figure*}

To elaborate, we start by extracting target entities involved in queries that require spatial relationship judgment. Then, utilizing object detection and scene graph generation algorithms, we gather geometric spatial position information and scene graph data of various entities from multi-modal visual inputs. This process involves employing an entity matching algorithm to obtain geometric spatial position information and scene graph details of the target entities. Finally, employing a corresponding prompt, we guide the large language model to respond to spatial awareness-related questions based on the geometric spatial position information and scene graph data of the target entities, generating the corresponding responses as depicted in Figure~\ref{fig:2}.

\subsection{Target Entity Extraction}
The ``REQUEST'' includes various elements, such as the user's ``Question'' and the input of multi-modal data. This paper initially extracts the target entities requiring determination of spatial relationships from the user's input ``Question,'' as illustrated in Formula 1:
\begin{equation}
\begin{split}
    (Entity_1, Entity_2) = \\Target~Entity~Extraction(Question)
    \label{eq:1}
\end{split}
\end{equation}
Here, $(Entity_1, Entity_2)$ represents the two target entities requiring relationship determination within the question.
The $Target~Entity~Extraction$ algorithm utilizes the $en\_core\_web\_sm$ model provided by the spaCy library. This model aims to offer lightweight natural language processing capabilities, including tokenization, part-of-speech tagging, named entity recognition, and dependency parsing. It demonstrates efficient performance in handling English textual data, making it well-suited for this task. Moreover, considering the involved multi-modal tasks typically involve the determination of relationships between two entities, the extraction process retrieves two target entities from the ``Question.''

\subsection{Geometric Spatial Location Information}
Geometric spatial position information refers to the geometric relative positioning details among objects in visual input. These specifics can be acquired through various algorithms, such as Object Detection, Stereo Vision, Depth Estimation, among others. This paper opts to employ Object Detection, an easily applicable algorithm in images, to obtain geometric spatial positioning details. Object Detection accurately locates the coordinates of objects in images, usually represented in the form of bounding boxes, which indicate the object's position and size. Additionally, the Object Detection algorithm conducts object classification and recognition, categorizing detected objects into predefined classes like humans, vehicles, animals, or items, facilitating the determination of object categories.

Specifically, we first use an object detection algorithm to obtain the position information $(x'_i,y'_i,w'_i,h'_i)$ and category $(E'_i)$ of objects in the image:
\begin{equation}
\begin{split}
        \{E'_1:(x'_1,y'_1,w'_1,h'_1),...,E'_m:(x'_m,y'
_m,w'_m,h'_m)\}\\=Object~Detection(Image)
\end{split}
\end{equation}
Here, $Image$ represents the input image, $\{E'_1:(x'_1,y'_1,w'_1,h'_1),...,E'_m:(x'_m,y'
_m,w'_m,h'_m)\}$ is a collection of detected entities' categories and corresponding geometric position coordinates, where m represents the number of detected entities, and the $Object~Detection$ algorithm employs Faster R-CNN~\cite{ren2017faster}.

Next, we match the detected entity categories $(E'_1,...,E'_m)$ from the visual input with the two target entities 
$(Entity_1, Entity_2)$ requiring a relationship from the "Question" to obtain a dictionary of position information closest to the entities $Entity_1$ and $Entity_2$ in the image: $\{Entity_1:(x_1,y_1,w_1,h_1), Entity_2:(x_2,y_2,w_2,h_2)\}$.

\subsection{Scene Graph Information}
Geometric spatial information is primarily concerned with identifying and locating the geometric relative position information among entities. However, in certain application scenarios, spatial awareness demands more than just understanding the geometric relative positions of entities. It also requires comprehension of higher-level semantic information among entities, such as semantic relationships between entities and semantic relationships between entities and the scene. This study utilizes Scene Graph Generation (SGG) algorithms to obtain corresponding scene graphs containing higher-level semantic information. By combining geometric spatial information from multi-modal visual inputs with scene graph details, a more comprehensive and accurate understanding of images is achieved. This method not only identifies the positions of objects but also understands their interactions within the scene. This is particularly beneficial in images with complex scenes and enhances the system's ability to address multi-modal tasks.

Specifically, we begin by utilizing a Scene Graph Generation (SGG) algorithm to generate the scene graph of the input image: \begin{equation}
    \{(s'_1,p'_1,o'_1),...,(s'_n,p'_n,o'_n)\}=SGG(Image)
\end{equation}
Here, $Image$ refers to the input image, and $\{(s'_1,p'_1,o'_1),...,(s'_n,p'_n,o'_n)\}$ represents the collection of triples obtained from the image scene graph, with $n$ indicating the number of triples. The $SGG$ algorithm is implemented using PSG~\cite{yang2022panoptic}.

Simultaneously, based on the collection of image scene graph triples, we extract all triples relevant to the target entities mentioned in the ``Question.'' NLTK library and the English lexical database WordNet are utilized for synonym matching. The matching criterion involves retaining a triple if one entity from the scene graph's triples matches $(Entity_1, Entity_2)$. Ultimately, this process generates a final set of target triples: $\{(s_1,p_1,o_1),...,(s_z,p_z,o_z)\}$, where $z$ denotes the number of resulting triples.

\subsection{Prompt Design}
Upon acquiring the geometric spatial information of the entities related to the question and the scene graph information, the crucial task is to effectively leverage this information to guide the MLLM in accurately answering spatial awareness-related user queries. Drawing from existing work~\cite{shao2023prompting, liu2023pre}, we have devised the following prompt format (refer to Table~\ref{tab:1}), which empowers the MLLM to utilize the spatial awareness information from the small model while accurately addressing user inquiries.
\begin{table}[h]
  \centering
  \begin{tabular}{|p{0.94\linewidth}|}
    \hline
The scene in the picture has the following relationship $\{(s_1,p_1,o_1),...,(s_z,p_z,o_z)\}$. And, The faster R-CNN detects the target and its geometric position as follows $\{Entity_1:(x_1,y_1,w_1,h_1), Entity_2:(x_2,y_2,w_2,h_2)\}$. Please answer the following questions based on the above information and the image itself: Question, and directly tell me the answer which you think is correct directly.\\
    \hline
   \end{tabular}
   \caption{The detail of the prompt design.}
   \label{tab:1}
\end{table}

\section{Experiments}
This section presents a comprehensive analysis of our proposed approach through a series of experiments. We start by outlining the Implementation Details, which offer key insights necessary for replicating the experiments and understanding the results. Next, we present the Main Results, showcasing the performance of our proposed model compared to existing large multi-modal models. Furthermore, we conduct ablation studies to validate the effectiveness of individual components in our proposed method.

\subsection{Implementation Details}
We provide a detailed description of the implementation aspects of our proposed approach. We outline the benchmarks used for experimentation, the baselines employed for comparison, the evaluation metrics used to assess model performance, and the hyperparameter settings of our experiments.

\textbf{Benchmarks.}
This paper primarily conducted experiments on two benchmarks, MME~\cite{fu2023mme} and MM-Vet~\cite{yu2023mm}, to validate the effectiveness of our approach. The MME benchmark consists of 10 perception tasks (existence, count, position, color, poster, celebrity, scene, landmark, artwork, OCR), where the position task specifically evaluates the model's spatial awareness capabilities. It comprises 957 images and 1914 QA pairs, with each image having two corresponding QA pairs. The MM-Vet benchmark includes 22 tasks, featuring 6 core tasks in computer vision and natural language processing (Recognition, Knowledge, OCR, Spatial Awareness, Language Generation, and Math), along with 16 combined tasks, where the spatial awareness task evaluates the model's spatial perception abilities. MM-Vet consists of 200 images and 218 questions paired with their respective ground truth answers, designed to cover diverse real-world scenarios with open-ended questions and expected answers.

\textbf{Baselines.}
This paper conducted experiments against a significant number of baselines to validate the effectiveness of our approach. The baselines used mainly included: BILP-2~\cite{li2023blip}, MiniGPT-4~\cite{zhu2023minigpt}, mPLUG-Owl~\cite{ye2023mplug}, ImageBind-LIM~\cite{han2023imagebind}, LLaMA-AdapterV2~\cite{gao2023llama}, VisualGLM-6B, Multimodal-GPT~\cite{gong2023multimodal}, PandaGPT~\cite{su2023pandagpt}, and LLaVA~\cite{liu2023visual}.

\textbf{Evaluation Metrics.}
This paper evaluates our model's performance using the evaluation metrics proposed in two benchmarks: MME and MM-Vet. Specifically, for the MME benchmark, the model's output is limited to two types (``yes'' or ``no''), making it convenient to measure accuracy and accuracy+ metrics. We choose to use the sum of accuracy and accuracy+ to calculate the task score. In the case of the MM-Vet benchmark, based on existing scoring instances and the model's output under the input question and real answer conditions for each sample, GPT-4 provides specific scores to evaluate the model's performance.

\textbf{Hyperparameter Settings.}
The hyperparameters used in this paper are the same as those in various baselines in MME and MM-Vet benchmarks. The key difference lies in our addition of geometric spatial position information and scene graph information into the model following the prompt format designed in this paper. Additionally, pre-trained target detection models and scene graph generation models were employed in this study.

\subsection{Main Results.}
We systematically tested the model's spatial awareness capabilities in the MME and MM-Vet benchmarks, and the experimental results are presented in Table~\ref{tab:2} and Table~\ref{tab:3}.

In the MME benchmark's ``position'' task, our model achieved an accuracy of 87.54, representing a 19.4\% improvement over the BLIP-2 base model. It also showed a notable 7.2\% improvement compared to the current leading MiniGPT-4 model, achieving a significant enhancement, surpassing the current state-of-the-art level.
For the spatial awareness task in the MM-Vet benchmark, our model attained an accuracy of 20.1, signifying a 24.1\% improvement over the BLIP-2 base model, showcasing significant performance enhancement.
\begin{table}
    \centering
    \begin{tabular}{lc}
        \hline
        Model & Position \\
        \hline
        ImageBind-LIM & 46.67\\
        LLaMA-AdapterV2 & 48.33\\
        VisualGLM-6B     & 48.33 \\
        PandaGPT  & 50.00\\
        mPLUG-Owl    & 50.00 \\
        LLaVA  & 50.00\\
        Multimodal-GPT & 58.33\\
        MiniGPT-4    & 81.67  \\
        BLIP-2-12B     & \textbf{73.33} \\
        Our Model & \textbf{87.54}\\
        \hline
    \end{tabular}
    \caption{The experimental results of the model in the position task of the MME benchmark.}
    \label{tab:2}
\end{table}

\begin{table}
    \centering
    \begin{tabular}{lc}
        \hline
        Model & Spatial Awareness \\
        \hline
        Transformers Agent (GPT-4) & 12.4 \\
        LLaMA-Adapter v2-7B  & 16.6 \\
        OpenFlamingo-9B & 18.0\\
        Otter-9B & 19.3\\
        InstructBLIP-8B & 18.6 \\
        BLIP-2-12B    & \textbf{16.2} \\
        Our Model & \textbf{20.1}\\
        \hline
    \end{tabular}
    \caption{The experimental results of the model in the spatial awareness task of the MM-Vet benchmark.}
    \label{tab:3}
\end{table}

\begin{table*}
    \centering
    \clearpage
    \begin{tabular}{lcccccccccc}
        \hline
         Model & Exist & Count & Position & Color & Poster & Celeb & Scene & Landm & Artwork & OCR  \\
         \hline
         MiniGPT-4 & 115.00 & 123.33 & 81.67 & 110.00 & 55.78 & 65.29 & 95.75 & 69.00 & 55.75 & 95.00\\
         mPLUG-Owl & 120.00 & 50.00 & 50.00 & 55.00 & 136.05 & 100.29 & 135.50 & {159.25} & 96.25 & 65.00\\
         ImageBind-LIM & 128.33 & 60.00 & 46.67 & 73.33 & 64.97 & 76.47 & 113.25 & 62.00 & 70.75 & 80.00\\
         VisualGLM-6B & 85.00 & 50.00 & 48.33 & 55.00 & 65.99 & 53.24 & 146.25 & 83.75 & 75.25 & 42.50\\
         Multimodal-GPT & 61.67 & 55.00 & 58.33 & 68.33 & 57.82 & 73.82 & 68.00 & 69.75 & 59.50 & 82.50\\
         PandaGPT & 70.00 & 50.00 & 50.00 & 50.00 & 76.53 & 57.06 & 118.00 & 69.75 & 51.25 & 50.00\\
         LLaVA & 50.00 & 50.00 & 50.00 & 55.00 & 50.00 & 48.82 & 50.00 & 50.00 & 49.00 & 50.00\\
         BLIP-2-12B & \textbf{160.00} & \textbf{135.00} & \textbf{73.33} & \textbf{148.33} & \textbf{141.84} & \textbf{105.59} & \textbf{145.25} & \textbf{138.00} & \textbf{136.50} & \textbf{110.00}\\
         Our Model & \textbf{168.00} & \textbf{135.00} & \textbf{86.67} & \textbf{145.00} & \textbf{141.84} & \textbf{105.59} & \textbf{147.98} & \textbf{137.25} & \textbf{135.50} & \textbf{110.00}\\
         \hline
    \end{tabular}
    \caption{The experimental results of the model on other tasks within the MME benchmark are as follows. Where Exist stands for Existence, Celeb represents Celebrity, and Landm indicates Landmarks.}
    \label{tab:4}
\end{table*}

\begin{table*}
    \centering
    \begin{tabular}{lccccccc}
        \hline
        Model & Rec & OCR & Know & Gen & Spat & Math & Total \\
        \hline
        Transformers Agent (GPT-4) & 18.2 & 3.9 & 2.2 & 3.2 & 12.4 & 4.0 & 13.4±0.5\\
        LLaMA-Adapter v2-7B     & 16.8 & 7.8 & 2.5 &  3.0 &  16.6 & 4.4 & 13.6±0.2 \\
        OpenFlamingo-9B & 24.6 & 14.4 & {13.0} & 12.3 & 18.0 & {15.0} & 21.8±0.1  \\
        MiniGPT-4-8B & 27.4 & {15.0} & 12.8 & {13.9} & {20.3} & 7.7 & 22.1±0.1 \\
        BLIP-2-12B &  \textbf{27.5} & \textbf{11.1} & \textbf{11.8} & \textbf{7.0} & \textbf{16.2} & \textbf{5.8} & \textbf{22.4±0.2}\\
        Our Model & \textbf{29.0} & \textbf{14.2} & \textbf{11.9} & \textbf{6.2} & \textbf{20.1} & \textbf{5.8} & \textbf{23.6±0.3}\\
        \hline
    \end{tabular}
    \caption{The experimental results of the model on the 6 core vision-language capabilities in the MM-Vet benchmark, where Rec, Know, Spat, and Gen represent recognition, knowledge, spatial awareness, and language generation, respectively, are presented.}
    \label{tab:5}
\end{table*}

\begin{table*}
    \centering
    \begin{tabular}{lccccccccccc}
        \hline
        Model & R/K/G & R & O/S & O/S/M & R/S & O & R/O/G/S& R/O/S & O/K/S \\
        \hline
        Transformers Agent(GPT-4) & 1.3 & 49.1 & 0.0 & 7.4 & 45.8 & 0.0 & 9.5 &  0.0 & 0.0\\
        LLaMA-Adapter v2-7B  & 0.2 & 43.2 & 7.9 & 8.1 & 41.7 & 0.0 & 26.8 & 0.0 & 33.3\\
        OpenFlamingo & 15.6 & 48.6 & 17.3 & 21.4 & 41.7 & 18.3 &0.0&14.3& 0.0 \\
        Otter-9B & 22.5 & 50.0 & 18.1 & 21.4 & 33.3 & 16.7 & 28.5 & 0.0 & 16.7 \\
        BLIP-2-12B &  \textbf{7.3} & \textbf{65.1} & \textbf{11.5} & \textbf{7.1} & \textbf{41.7} & \textbf{21.2} & \textbf{8.5} & \textbf{14.3} & \textbf{16.7}\\
        Our Model & \textbf{8.1} & \textbf{67.3} & \textbf{15.4} & \textbf{8.1} & \textbf{47.5} & \textbf{22.4} &  \textbf{12.8} & \textbf{18.3} & \textbf{33.3}\\
        \hline
    \end{tabular}
    \caption{The experimental results of the model on the 9 integrations of interest derived from the capability combination in the MM-Vet benchmark. Where R, K, G, O, S, and M represent Recognition, Knowledge Language, Generation, OCR, Spatial Awareness, and Math, respectively.}
    \label{tab:6}
\end{table*}

In addition to evaluating the performance of our method on two benchmarks specifically designed to assess spatial awareness, this study also tested the effectiveness of our method on other tasks. In the MME benchmark, our method not only significantly improved in the position task but also exhibited enhancements in other related tasks, as shown in Table~\ref{tab:4}. Specifically, in the existence task, our method raised the model's accuracy from the baseline model BLIP-2's 160.00 to 168.00, marking a 5\% improvement. In the scene task, our method raised the model's accuracy from the baseline model BLIP-2's 145.25 to 147.98, a 1.9\% increase, both achieving the current best levels. Moreover, the experimental results indicate that our model generally maintained the performance of the baseline model in tasks unrelated to spatial awareness.

On the MM-Vet benchmark, our approach not only improved the model's performance in spatial awareness tasks but also enhanced its performance in other related core vision-language tasks, as depicted in Table~\ref{tab:5}. Specifically, in the recognition task, our method increased the model's accuracy from 27.5 to 29.0, marking a 5.5\% improvement. For the OCR task, our method raised the model's accuracy from 11.1 to 14.1, showing a significant improvement of 27.9\%. Moreover, in the associated mixed tasks, our method also notably enhanced the model's performance. As demonstrated in Table~\ref{tab:6}, among the 16 mixed tasks, our method outperformed the base model in 10 tasks. For instance, in the Recognition/Knowledge/Language Generation task, the model's accuracy increased by 11.0\%; in the OCR/Spatial Awareness task, the model's accuracy improved by 3.9\%; in the OCR/Spatial Awareness/Math task, the model's accuracy rose by 14.1\%; and in the OCR/Knowledge/Spatial Awareness task, the model's accuracy surged by 99.4\%.

In summary, our approach not only significantly enhances the model's spatial awareness capabilities but also markedly improves the model's performance in other aspects such as object recognition (i.e., existence and recognition tasks), scene understanding (i.e., scene and OCR tasks), and mixed tasks. Consequently, our approach can more effectively enhance the overall performance of multimodal large models, demonstrating considerable and broad effectiveness.

\subsection{Ablation Studies.}
We also conducted extensive ablation experiments on both the MME and MM-Vet benchmarks to verify the effectiveness of the proposed geometric spatial position information, scene graph information, and their fusion in enhancing spatial awareness capabilities. The ablation experiment prompts are presented in Table~\ref{tab:7}.

\begin{table*}
    \centering
    \begin{tabular}{p{2.8cm}|p{13.9cm}}
        \hline
        Model & \multicolumn{1}{c}{Prompt Detail} \\
        \hline
        BLIP-2+GPL & The Faster R-CNN detects the object and its geometric position as follows $\{Entity_1:(x_1,y_1,w_1,h_1), Entity_2:(x_2,y_2,w_2,h_2)\}$. Please answer the question based on the image and geometric position of the object: Question, and directly tell me the answer which you think is correct directly.\\
        \hline
        BLIP-2+SG & The scene in the picture has the following relationship $\{(s_1,p_1,o_1),...,(s_z,p_z,o_z)\}$. Please answer based on the scene relationship and the image: Question And tell me the answer which you think is correct directly.\\
        \hline
        BLIP-2+SG+GPL & The scene in the picture has the following relationship$\{(s_1,p_1,o_1),...,(s_z,p_z,o_z)\}$. And, The Faster R-CNN detects the target and its geometric position as follows $\{Entity_1:(x_1,y_1,w_1,h_1),Entity_2:(x_2,y_2,w_2,h_2)\}$. Please answer the following questions based on the above information and the image itself: Question, and directly tell me the answer which you think is correct directly.\\
        \hline
    \end{tabular}
    \caption{Prompt design in ablation experiments. Where GPL represents geometric position location, SG stands for scene graph.}
    \label{tab:7}
\end{table*}

The results of the ablation experiments on the MME benchmark are shown in Table~\ref{tab:8}. The experimental results reveal that the proposed geometric spatial position information enhanced the model's accuracy in the position task from 73.33 to 78.36, demonstrating a 6.9\% improvement, highlighting the effectiveness of geometric spatial position information. Likewise, the introduced scene graph information increased the accuracy of the model in the position task from 73.33 to 80.48, displaying a 9.6\% improvement, showcasing the effectiveness of scene graph information. Most significantly, the combined use of geometric spatial position information and scene graph information enhanced the model's accuracy in the position task from 73.33 to 87.54, showcasing a 19.4\% improvement, emphasizing the effectiveness of combining information in advancing spatial awareness in large multimodal language models. Additionally, the improvement from the fusion method was 2.9\% higher than the cumulative improvement from using the two pieces of information separately, effectively demonstrating the advantage of integrating information from scene graph generation algorithms and object detection algorithms in better understanding the positioning relationships between target entities in an image, ultimately enhancing the performance and effectiveness of multimodal tasks.
\begin{table}
    \centering
    \begin{tabular}{lc}
        \hline
        Model & Position \\
        \hline
        BLIP-2-12B     & 73.33 \\
        BLIP-2-12B+GPL & \textbf{78.36(+6.9\%)}\\
        BLIP-2-12B+SG & \textbf{80.48(+9.6\%)}\\
        BLIP-2-12B+GPL+SG (Our Model) & \textbf{87.54(+19.4\%)}\\
        \hline
    \end{tabular}
    \caption{The model's ablation experiments for the position task on the MME benchmark. Where GPL represents geometric position location, SG stands for scene graph.}
    \label{tab:8}
\end{table}

The ablation experiment results on the MM-Vet benchmark are shown in Table~\ref{tab:9}. From the experimental findings, it is evident that the proposed geometric spatial position information increased the model's accuracy in the spatial awareness task from 16.2 to 18.8, showcasing a 16.0\% enhancement, highlighting the effectiveness of geometric spatial position information. Similarly, the introduced scene graph information enhanced the model's accuracy in the spatial awareness task from 16.2 to 19.6, demonstrating a 21.0\% improvement, underlining the effectiveness of scene graph information. Most significantly, the combined utilization of geometric spatial position information and scene graph information elevated the model's accuracy in the spatial awareness task from 16.2 to 20.1, displaying a 24.1\% enhancement, emphasizing the effectiveness of combining information to improve spatial awareness in large multimodal language models. It also attests to the widespread effectiveness of the proposed method across different datasets.
\begin{table}
    \centering
    \begin{tabular}{lc}
        \hline
        Model & Spatial Awareness \\
        \hline
        BLIP-2-12B     & 16.2 \\
        BLIP-2-12B+GPL & \textbf{18.8(+16.0\%)}\\
        BLIP-2-12B+SG & \textbf{19.6(+21.0\%)}\\
        BLIP-2-12B+GPL+SG (Our Model) & \textbf{20.1(+24.1\%)}\\
        \hline
    \end{tabular}
    \caption{The model's ablation experiments for the spatial awareness task on the MM-Vet benchmark. Where GPL represents geometric position location, SG stands for scene graph.}
    \label{tab:9}
\end{table}

\section{Conclusion}
The paper introduces a novel approach to enhance the spatial awareness capabilities of MLLM. This method utilizes pre-trained geometric spatial information extraction algorithms and scene graph generation algorithms to provide geometric spatial information between objects and higher-level scene graph details. This guidance aids MLLM in more accurately addressing user inquiries related to spatial awareness. Extensive experiments were conducted on benchmarks such as MME, MM-Vet, and other MLLM benchmarks. The experimental outcomes robustly confirm the efficacy of this research method, significantly improving MLLM performance in spatial awareness and associated tasks.

\bibliography{aaai24}

\end{document}